\documentclass{article}

\PassOptionsToPackage{numbers, sort&compress}{natbib}

     \usepackage[preprint, nonatbib]{neurips_2024}

\usepackage[utf8]{inputenc} 
\usepackage[T1]{fontenc}    
\usepackage{hyperref}       
\usepackage{url}            
\usepackage{booktabs}       
\usepackage{amsfonts}       
\usepackage{nicefrac}       
\usepackage{microtype}      
\usepackage{xcolor}         
\usepackage{amsbsy}
\usepackage{amsmath}
\usepackage[square,numbers]{natbib}
\usepackage{graphicx}
\usepackage{nicematrix}
\usepackage{pifont}%
\usepackage{subcaption}
\usepackage{multirow}

\newcommand{\cmark}{\ding{51}}%
\newcommand{\xmark}{\ding{55}}%

\newcommand{\methodName}{\textsc{CountGD}}

\title{\methodName: Multi-Modal Open-World Counting}

\author{%
 Niki Amini-Naieni \qquad  Tengda Han  \qquad Andrew Zisserman\\
 Visual Geometry Group (VGG)\\
 University of Oxford\\
 \texttt{\{nikian,htd,az\}@robots.ox.ac.uk} 
}

\begin{document}

\maketitle

\begin{abstract}
The goal of this paper is to improve the generality and accuracy of open-vocabulary object counting in images.
To improve the generality, we repurpose an open-vocabulary detection  foundation model (GroundingDINO) 
for the counting task, and also extend its capabilities by introducing modules to enable specifying the target object to count by visual exemplars.
In turn, these new capabilities -- being able to specify the target object by multi-modalites (text and exemplars) -- lead to an improvement in counting accuracy.

We make three contributions:
{\em first}, we introduce the first open-world counting model, \methodName,  where the prompt can be specified by a text description or visual exemplars or both;
{\em second}, we show that the performance of the model significantly
improves the state of the art on multiple counting benchmarks -- 
 when using text only, \methodName\ is comparable to or outperforms all previous text-only works, and  when using both text
and visual exemplars, we outperform all previous models;
{\em third}, we carry out a preliminary study into different interactions between the text and visual exemplar prompts, including the cases where they reinforce each other and where one restricts the other. The code and an app to test the model are available at \href{https://www.robots.ox.ac.uk/~vgg/research/countgd/}{https://www.robots.ox.ac.uk/~vgg/research/countgd/}.

\end{abstract}


\begin{figure}[h!]
\includegraphics[width=\linewidth]{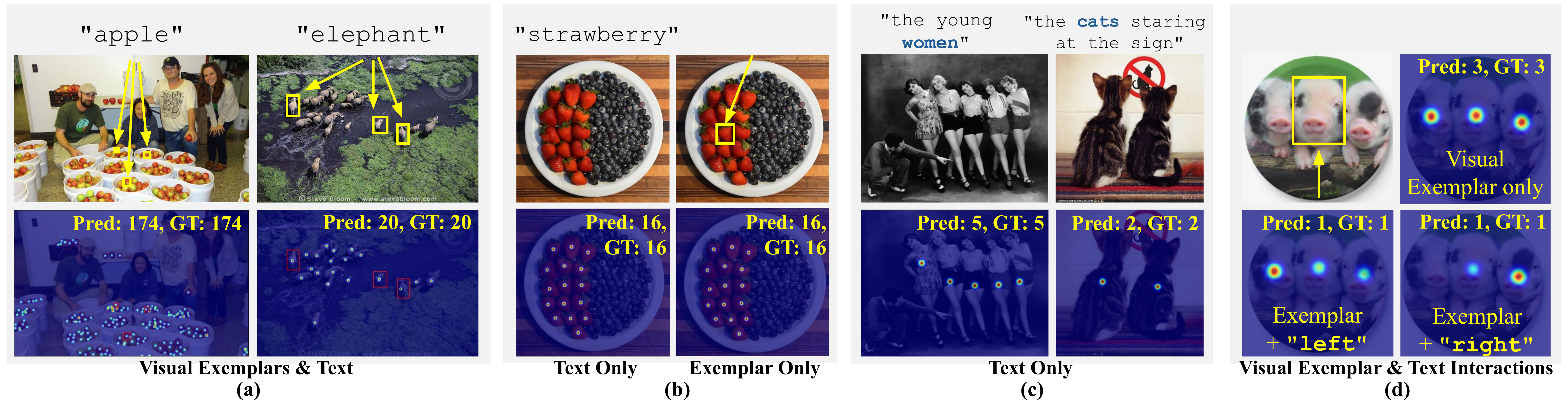}
\vspace{-6mm}
\caption{\small 
\methodName~ is capable of taking \emph{both} visual exemplars and text prompts to produce highly accurate object counts \textbf{(a)},
but also seamlessly supports counting with only text queries or only visual exemplars \textbf{(b)}.
The multi-modal visual exemplar and text queries bring extra flexibility to the open-world counting task, such as using a short phrase \textbf{(c)}, or adding additional constraints (the words `left' or `right') to select a sub-set of the objects \textbf{(d)}.
These examples are taken from the FSC-147~\cite{m_Ranjan-etal-CVPR21} and CountBench~\cite{paiss2023countclip} test sets.
The visual exemplars are shown as yellow boxes.
(d) visualizes the predicted confidence map of the model, where a high color intensity indicates a high level of confidence.
}
\vspace{-2mm}
\label{fig:teaser}
\end{figure}

\vspace{-2mm}
\section{Introduction}
\vspace{-2mm}

Open-world object counting methods aim to enumerate all the instances
of any category of object in an image. The `open-world' refers to the
model's ability to count objects beyond the set of categories seen at training, 
thus enabling the user to specify categories of interest at inference without the need for model retraining.
Recent techniques allow the user to specify the target object with only visual exemplars --
bounding boxes around a few example objects in the image --~\cite{Lu18, Liu22}, or
only text descriptions~\cite{AminiNaieni23, Jiang2023CLIPCountTT}. By accepting either visual exemplars or
text as {\em prompts}, open-world object counting methods can adapt to the specific object at inference time. This
enables these techniques to count arbitrary classes of objects as specified by the user.

Methods that use visual exemplars to specify the object currently significantly outperform text-based
counting methods on multiple benchmarks. This is because visual
exemplars provide more detailed information than text -- it can take many words to precisely describe an object; and 
perhaps more importantly, they provide {\em intrinsic} information on the object's appearance -- 
because the exemplars are from the same image they already `factor in' the viewpoint and lighting, variables that significantly affect the object's appearance.
However, while visual
exemplar-based approaches are more accurate,  they limit the capabilities and generality of the counting model.

In this paper we introduce a counting model that is able to specify
the target object using visual exemplars, a text description, or both
together. The model, named \methodName, has superior accuracy to
previous methods, but is also more
general. In addition to the performance boost obtained by specifying the target object
using both visual exemplars and text, the {\em interaction} of the exemplars and text can be used
to select a sub-set of those objects in the image. These capabilities are illustrated in Figure~\ref{fig:teaser}.
This flexible
combination of visual exemplars and text description thus provides the
model with more capabilities and information than prior approaches.

To achieve this multi-modal prompt capability, we 
follow prior work on
open-world text-specified object counting~\cite{AminiNaieni23}, and build on and extend
a pre-trained vision-language foundation model,  GroundingDINO~\cite{liu2023grounding}. 
We introduce new modules to embed the visual exemplars, and to enable the model to count, rather than detect.
Within the model we cast the additional visual exemplars as text tokens, and 
the model first learns to fuse the visual exemplars with text tokens through self-attention, and then interacts
with the image through cross-attention. 
Because the text tokens
are naturally variable in length, the number of provided visual
exemplars are as well. As a result, the model allows users to specify
the object to count with text only, visual exemplars only, or text and
any number of visual exemplars.

In summary, we make the following three contributions:
\emph{First}, we introduce \methodName, the first open-world object counting model that
accepts either text or visual exemplars or both simultaneously, in a single-stage architecture;
\emph{Second}, we evaluate the model on multiple standard counting benchmarks, including FSC-147~\cite{m_Ranjan-etal-CVPR21}, CARPK~\cite{Hsieh2017DroneBasedOC} and CountBench~\cite{paiss2023countclip}, and show
that \methodName\ significantly improves on the
state-of-the-art performance by specifying the target object using both exemplars and text. It also meets or improves on the state-of-the-art for text-only approaches when trained and evaluated using text-only; 
\emph{Third}, we investigate how the text can be used to refine the visual information provided by the exemplar, for example by filtering on color or relative position in the image, to specify a sub-set of the objects to count.
In addition we make two minor improvements to the inference stage: one that addresses the problem of double counting
due to self-similarity, and the other to handle the problem of a very high count.

\vspace{-2mm}
\section{Related Work}
\vspace{-2mm}

Prior work on object counting has developed along three axes: (1) the density map versus detection axis, (2) the class-specific versus open-world (also referred to as ``class-agnostic") axis, and (3) the visual exemplar versus text axis. The pattern is that detection, open-world, and text-based methods tend to offer more capabilities and be more general than their analogues along each axis. On the other hand, density map, class-specific, and visual exemplar-based methods tend to be more accurate at the counting tasks they apply to. \methodName\ integrates the third axis -- the visual exemplar versus text axis -- to achieve more general and accurate counting overall. Below, we discuss where prior work falls along each axis and where \methodName\ stands.

\vspace{-4mm}
\paragraph{Density Map versus Detection-based Object Counting \emph{(Axis 1)}.}
In the past, counting techniques that regress and sum density maps~\cite{Arteta14, Arteta16, Cho1999ANC, Kong2006AVI, Lempitsky10b, Marana1997EstimationOC, Xie16}, instead of detecting and enumerating bounding boxes~\cite{Barinova2010OnDO, 10.1007/s11263-011-0439-x, Hsieh2017DroneBasedOC, 10.1007/978-3-031-20044-1_20}, have proven more accurate in cluttered and dense scenes.
For example, density map-based approaches like CounTX~\cite{AminiNaieni23}, LOCA~\cite{low_shot}, and CounTR~\cite{Liu22} achieve lower counting errors than detection-based approaches such as Mask-RCNN~\cite{mask-rcnn} and RetinaNet \cite{retina-net} on standard counting benchmarks. Concurrent to our work, DAVE~\cite{dave}, integrates density map regression with object detection to construct a more accurate and explainable two-stage counting system. Like DAVE, \methodName\ outputs explicit object locations. However, \methodName\ is a single-stage approach that achieves better counting accuracy than DAVE and other density map-based techniques. Therefore, while density map-based approaches tend to be more accurate than detectors in highly populated scenes, recent detection-based techniques, including \methodName, are beginning to achieve better accuracy than density map-based alternatives.

\vspace{-3mm}
\paragraph{Class-specific versus Open-world Object Counting \emph{(Axis 2)}.}
Object counting methods first developed as class-specific techniques~\cite{Arteta16, 10.1007/978-3-031-19821-2_11, mundhenkLargeContextualDataset2016a, Xie16}, solving the counting problem for only one category of object, but recent methods have generalized these approaches to open-world settings, where counting \emph{arbitrary} objects is possible. Class-specific methods have been developed to count cars~\cite{Kili2021AnAC}, humans~\cite{10.1007/978-3-031-19821-2_11}, and cells~\cite{Flaccavento11}. In contrast, open-world methods can count instances from all three categories~\cite{Lu18}. Because class-specific techniques are more specialized than open-world approaches, they tend to be more accurate at counting instances from the class they were designed for. 
{Recent advancements in Vision-Language Foundation Models (VLMs) such as CLIP~\cite{Radford2021LearningTV} and GroundingDINO~\cite{liu2023grounding} trained on web-scale image-text pairs produce semantically rich visual and textual features. These features generalize to a wide range of open-world downstream tasks.
Building on top of these pre-trained VLMs}, recent open-world methods~\cite{groundingrec, AminiNaieni23, kang2024vlcounter, Shi2022RepresentCA, Liu22, low_shot, You_2023_WACV} have begun to surpass class-specific approaches in counting accuracy.
\methodName, like these recent approaches, is an open-world object counter that achieves competitive performance in comparison to class-specific alternatives.

\vspace{-3mm}
\paragraph{Counting with Visual Exemplars versus Counting with Text \emph{(Axis 3)}.}
Most open-world object counters approach the problem by using visual exemplars to select the objects in the input image~\cite{Gong2022ClassAgnosticOC, Liu22, Lu18, 10.1007/978-3-031-20044-1_20, m_Ranjan-etal-CVPR21, Shi2022RepresentCA, yangClassagnosticFewshotObject2021,  You_2023_WACV, low_shot, Lin_2022_BMVC}, but very recent work~\cite{groundingrec, AminiNaieni23, Jiang2023CLIPCountTT, Xu2023ZeroshotOC, kang2024vlcounter} has attempted to replace the visual exemplars with text, enabling new capabilities at the cost of reduced accuracy. The state-of-the-art text-based approaches, such as GroundingREC~\cite{groundingrec}, CounTX~\cite{AminiNaieni23}, CLIP-Count~\cite{Jiang2023CLIPCountTT}, and VLCounter~\cite{kang2024vlcounter} are built on top of vision-language foundation models pretrained on large quantities of data to relate images to textual inputs and map them to a joint embedding space. This allows these foundation models to understand general concepts learned during extensive pretraining and provides a mechanism for users to specify extrinsic object properties through text. 
However, text-based approaches perform significantly worse than state-of-the-art visual exemplar-based approaches such as LOCA~\cite{low_shot}, CounTR~\cite{Liu22}, and few-shot DAVE~\cite{dave}. For example, while both GroundingREC and \methodName\ use the pretrained GroundingDINO~\cite{liu2023grounding} vision-language foundation model, unlike GroundingREC, \methodName\ allows the user to input both visual exemplars and text instead of just text. This enables \methodName\ to achieve superior counting accuracy in comparison to GroundingREC.
Notably, DAVE~\cite{dave} is a visual exemplar-based approach that also enables textual prompts, but differs from \methodName\ in three important ways: (1) it does not address the case when both text and visual exemplars are available while \methodName\ does, 
(2) its comparison between text features and image features is not learned as it is by \methodName\ with attention, and (3) it is a two-stage approach, while \methodName\ solves the problem in a single stage, without relying on another visual exemplar-based counting model. 

\vspace{-2mm}
\paragraph{Relation of Counting to other areas.}
Our work is related to few-shot image classification~\cite{vinyals2016matching} and image detection~\cite{kang2019few,fan2020few} methods. 
These works require a few query images of novel objects, and then compare the test image with these image examples to determine its semantic content (for image classification), or to spatially localize instances (for object detection). Like these methods, \methodName\ enables us to specify the object to count with visual exemplars (i.e., ``query images") but also allows for textual inputs, and then compares the test image with the multi-modal specifications to get the final count. Furthermore, we focus on the counting problem, a challenging task for object detectors. 


\vspace{-5mm}
\section{Counting with Visual Exemplars \& Text} 
\vspace{-3mm}

Here, we describe
\methodName, a single-stage model for open-world object counting
that accepts either visual exemplars or text or both together as prompts to specify the object to count.


\subsection{Overview}\label{overview}
\vspace{-2mm}

Given a target object specified by either visual
exemplars as bounding boxes $\mathbf{B} = \{b_{1}, \cdots, b_{N}\}$
around example object instances in the image, or a textual
description, $t$, or both, $\{\mathbf{B}, t\}$, the 
counting model, $f$, counts the number of occurrences of the object in an image $\mathbf{X} \in \mathbb{R}^{H \times W \times 3}$,  as $\hat{y} = f(\mathbf{X}, \mathbf{B}, t)$,
where $\hat{y}$ is the object count estimated by the counting model $f$.


\begin{figure}[h!]
\centering
\includegraphics[width=\linewidth]{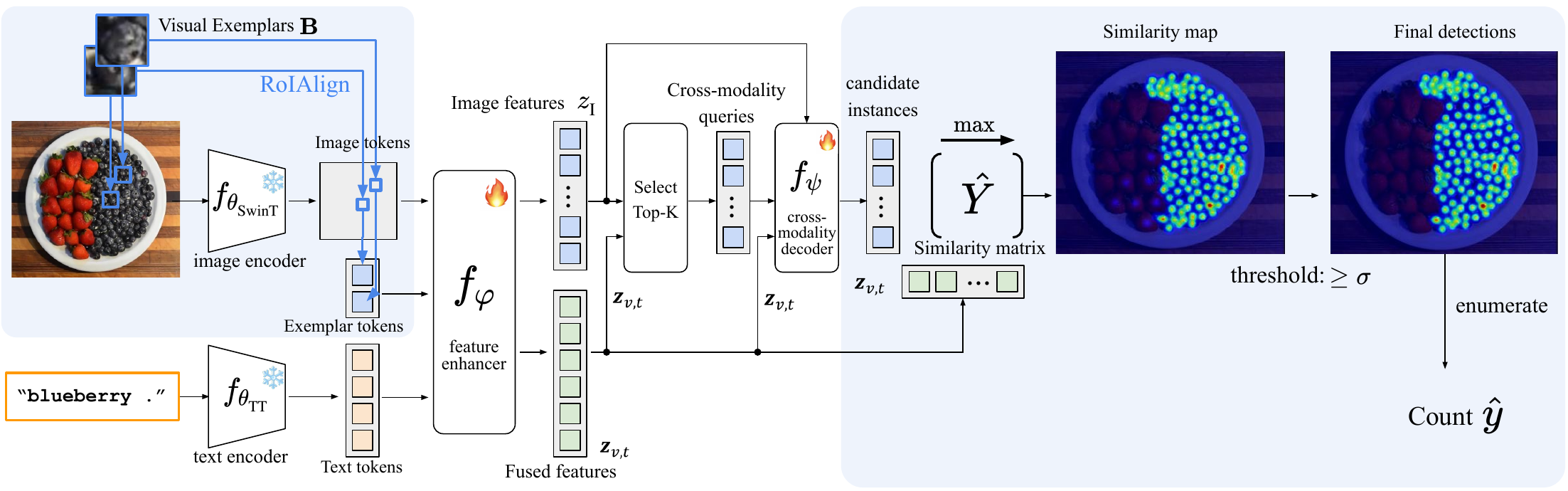}
\vspace{-6mm}
\caption{
\small The \methodName\ architecture. At inference the object to be counted can be specified by visual exemplars or text prompts or both. The input image is passed through the image encoder, $f_{\boldsymbol{\theta_\text{SwinT}}}$ to obtain spatial feature maps at different scales. The visual exemplar tokens are cropped out of this feature map using RoIAlign (as shown in Figure~\ref{fig:visual_encoder}). The text is passed through the text encoder, $f_{\boldsymbol{\theta_\text{TT}}}$ to obtain text tokens. In the feature enhancer, $f_{\boldsymbol{\varphi}}$, the visual exemplar tokens and text tokens are fused together with self-attention and cross-attend to the image features, producing the fused visual exemplar and text features, $\mathbf{z_{v, t}}$, and new image features, $\mathbf{z_{I}}$. The $k$ image features $\mathbf{z_{I}}$ that have the highest cosine similarity with the fused features $\mathbf{z_{v, t}}$ are passed to the cross-modality decoder, $f_{\boldsymbol{\psi}}$, as ``cross-modality queries". Finally, the similarity matrix, $\mathbf{\hat{Y}}$ between the outputs of the cross-modality decoder,  $f_{\boldsymbol{\psi}}$, and  $\mathbf{z_{v, t}}$ is calculated, and outputs that achieve a maximum similarity with the $\mathbf{z_{v, t}}$ above a confidence threshold $\sigma$ are identified as final detections and enumerated to estimate the final count. 
Our model is built on top of GroundingDINO~\cite{liu2023grounding} architecture with the additional modules indicated by blue shading.
}
\vspace{-2mm}
\label{architecture_fig}
\end{figure}

The architecture of the model is illustrated in Figure~\ref{architecture_fig}.
\methodName\ is built on top of the
open-world object detector GroundingDINO~\cite{liu2023grounding} to benefit from its pretrained open-vocabulary grounding and detection capabilities. 
In contrast to GroundingDINO,  which only uses text queries for object detection, 
\methodName\ also includes visual exemplars as inputs, 
which  increases the performance and flexibility of the model for object counting.
In the following, we first describe the modules of the \methodName\ architecture, and then discuss its relation to GroundingDINO and in particular what is frozen,  what is trained, and what is added to GroundingDINO.

\subsection{\methodName\ Architecture Components}\label{architecture}


\begin{figure}[t]
    \centering
    \includegraphics[width=\textwidth]{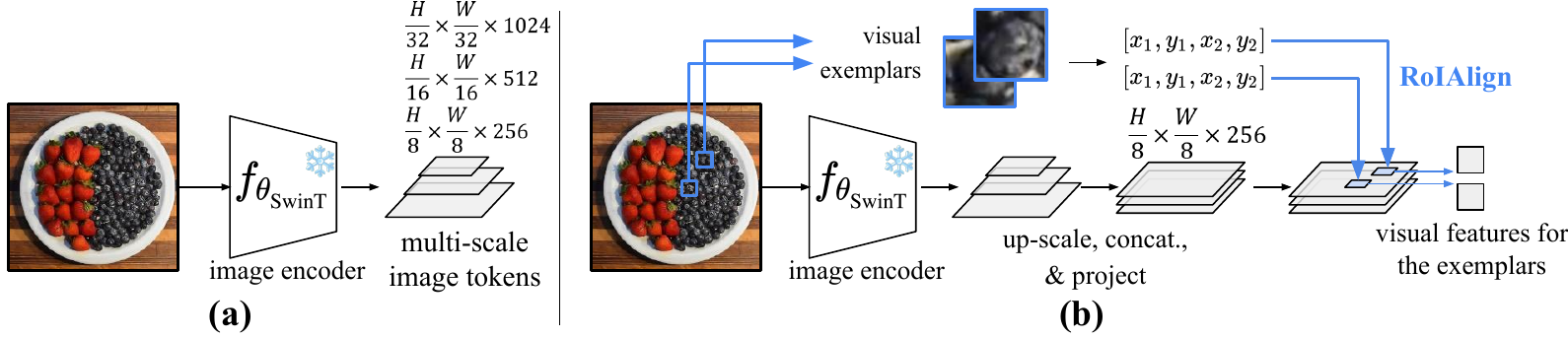}
    \vspace{-6mm}
    \caption{\small The visual feature extraction pipeline for images and visual exemplars. \textbf{(a)} For the input image, a standard Swin Transformer model is used to extract visual feature maps at multiple spatial resolutions. 
    \textbf{(b)} For the visual exemplars with their corresponding bounding boxes, we first up-scale the multiple visual feature maps of the input image to the same resolution, then concatenate these feature maps, and project them to 256 channels with a $1\times 1$ convolution. Finally, we apply a RoIAlign with the bounding box coordinates to get the visual features for the exemplars. 
    }
    \label{fig:visual_encoder}
    \vspace{-5mm}
\end{figure}

\paragraph{Image Encoder ($f_{\boldsymbol{\theta_{\text{SwinT}}}}$).}
The image encoder $f_{\boldsymbol{\theta_{\text{SwinT}}}}$ encodes two types of inputs: the input image $X$ and the visual exemplars $\mathbf{B}$.
The image encoder itself is the Swin-B version of the Swin Transformer~\cite{Liu2021SwinTH}.
As shown in Figure~\ref{fig:visual_encoder} (a), for the input image $X$,
it produces spatial feature maps at three different scales.
These spatial feature maps are projected to 256 dimensions with 1x1 convolutions to produce the image tokens, feature vectors of length $256$ corresponding to the image patches at different scales, which are input to the feature enhancer, $f_{\boldsymbol{\varphi}}$.
As shown in Figure~\ref{fig:visual_encoder} (b), for the visual exemplars $\mathbf{B}$, we reuse the spatial feature map $f_{\boldsymbol{\theta_{\text{SwinT}}}}(\mathbf{X})$ for the input image $X$, and apply aligned region-of-interest pooling, RoIAlign~\cite{he2017maskrcnn}, with the pixel coordinates specified by the visual exemplars $\mathbf{B}$. 
The resulting visual exemplar tokens are 256-dimensional feature vectors like the image and text tokens.

\vspace{-2mm}
\paragraph{Text Encoder ($f_{\boldsymbol{\theta_{\text{TT}}}}$).}
For the text encoder, $f_{\boldsymbol{\theta_{\text{TT}}}}$, we use the BERT-base~\cite{bert} text transformer pretrained on detection and phrase grounding data with the image encoder, $f_{\boldsymbol{\theta_{\text{SwinT}}}}$. The text encoder maps the input object description $t$ to a sequence of at most 256 tokens. The encoded text tokens are 256-dimensional feature vectors. While the image encoder $f_{\boldsymbol{\theta_{\text{SwinT}}}}$ produces $n$ image patch features when there are $n$ multi-scale patches extracted from the input image, and the visual exemplar encoder produces $p$ visual exemplar features when $p$ visual exemplars are available, the text encoder produces $q$ text features when there are $q$ tokens, as determined by the BERT tokenizer, in the text $t$. The $n$ image tokens, $p$ visual exemplar tokens, and $q$ text tokens are then passed to the feature enhancer $f_{\boldsymbol{\varphi}}$, which fuses the three sources of information with attention.

\vspace{-2mm}
\paragraph{Feature Enhancer ($f_{\boldsymbol{\varphi}}$).}
The feature enhancer, $f_{\boldsymbol{\varphi}}$, is composed of 6 blocks that first fuse the visual exemplar tokens with the text tokens through self-attention and then fuse the combined features with the image patch tokens with cross-attention. More specifically, each block consists of self-attention between the concatenated visual exemplar and text tokens, deformable self-attention between the image patch tokens, and image-to-text cross-attention and text-to-image cross-attention between the fused visual exemplar and text tokens and the image patch tokens. These modules enable \methodName\ to learn to relate information from 
the input image, visual exemplars and text query altogether.
The feature enhancer $f_{\boldsymbol{\varphi}}$ outputs two sets of
features denoted as $\mathbf{z_{v, t}}$ and $\mathbf{z_I}$ as
\begin{equation}\label{arch_eq_1}
    (\mathbf{z_{v, t}}, \mathbf{z_I}) = f_{\boldsymbol{\varphi}}\left(f_{\boldsymbol{\theta_{\text{SwinT}}}}(\mathbf{X}), \text{RoIAlign}(f_{\boldsymbol{\theta_{\text{SwinT}}}}(\mathbf{X}), \mathbf{B}), f_{\theta_{\text{TT}}}(t)\right)
\end{equation}
corresponding to the fused
visual exemplar and text tokens, and the image patch tokens,
respectively.

\vspace{-2mm}
\paragraph{Language \& Visual Exemplar-guided Query Selection ($Select$).}
We select the $k$ image patch tokens $\mathbf{z_I}$ that achieve the
highest similarity with the fused visual exemplar and text
tokens $\mathbf{z_{v, t}}$. This operation is denoted by
$\text{Select}\left(\mathbf{z_I}, \mathbf{z_I} \mathbf{z_{v, t}}^{T},
k\right)$,  where $\mathbf{z_I}
\mathbf{z_{v, t}}^{T} \in \mathbb{R}^{n \times (p + q)}$ represents
the similarity scores between the $n$ image patch tokens and the $p +
q$ visual exemplar and text tokens. As in
GroundingDINO~\cite{liu2023grounding}, we set $k$ to 900. 
These 900 image patch tokens with higher similarity scores
serve as ``cross-modality queries" input
to the cross-modality decoder $f_{\boldsymbol{\psi}}$.

\vspace{-2mm}
\paragraph{Cross-modality Decoder ($f_{\boldsymbol{\psi}}$).}
The cross-modality decoder, $f_{\boldsymbol{\psi}}$, uses
self-attention to enhance the cross-modality queries, image
cross-attention to fuse the image patch features $\mathbf{z_I}$ to the
cross-modality queries, and cross-attention to fuse the visual
exemplar and text features $\mathbf{z_{v, t}}$ to the cross-modality
queries. In more detail, the cross-modality decoder consists of 6 of
these self-attention and cross-attention blocks. 
The cross-modality queries are dot-producted with
the combined visual exemplar and text tokens $\mathbf{z_{v, t}}$ and
passed through an element-wise Sigmoid function  to obtain the final confidence scores as:
\begin{equation}\label{arch_eq_2}
\mathbf{\hat{Y}} = \text{Sigmoid}\left(f_{\boldsymbol{\psi}}\left(\mathbf{z_I}, \mathbf{z_{v, t}}, \text{Select}(\mathbf{z_I}, \mathbf{z_I} \mathbf{z_{v, t}}^{T}, k)\right) \mathbf{z_{v, t}}^{T}\right)
\end{equation}
where $\mathbf{z_{v, t}}$ are the fused visual exemplar and text
features, $\mathbf{z_I}$ are the image features, $k$ is the number of
queries (i.e., maximum number of detected objects), and
$\mathbf{\hat{Y}}$ are the final similarity scores that are
thresholded according to a confidence threshold $\sigma$ and
enumerated to estimate the final object count $\hat{y}$ at inference.

\vspace{-2mm}
\paragraph{Design choices and relation to GroundingDINO.}
We choose GroundingDINO~\cite{liu2023grounding} over other VLMs due to its pretraining on visual grounding data, providing it with more fine-grained features in comparison to other VLMs such as CLIP~\cite{Hobley2022LearningTC}. 


To extend GroundingDINO to accept visual exemplars, we cast them as
text tokens. Because both the visual exemplars and the text specify the object, we posit
that the visual exemplars can be treated in the same way as the text
tokens by GroundingDINO and integrate them into the training and
inference procedures as such. In treating the visual exemplars as additional text tokens within a phrase, we add self-attention between the phrase corresponding to the visual exemplar and the visual exemplar rather than keeping them separate. This allows \methodName\ to learn to fuse the visual exemplar and text tokens to form a more informative specification of the object to count.
Similarly, cross-attention between the image and text features
in GroundingDINO's feature enhancer and cross-modality decoder becomes
cross-attention between the image and the fused visual exemplar and
text features in \methodName. Language-guided query selection in
GroundingDINO becomes language and visual exemplar-guided query
selection in \methodName. In this way, \methodName\ naturally extends
GroundingDINO to input both text and visual exemplars to describe the
object.

In GroundingDINO, the image encoder $f_{\boldsymbol{\theta_{\text{SwinT}}}}$ is pre-trained on abundant detection and phrase grounding data with the text encoder, $f_{\boldsymbol{\theta_{\text{TT}}}}$, providing it with rich region and text-aware features. Since we wish to build on this pre-trained joint vision-language embedding, we keep the image encoder $f_{\boldsymbol{\theta_{\text{SwinT}}}}$ and the text encoder $f_{\boldsymbol{\theta_{\text{TT}}}}$ frozen.

\vspace{-2mm}
\subsection{Training}\label{training}
\vspace{-2mm}

We train the projection layers for extracting the visual exemplar tokens, the feature enhancer, and the cross-modality decoder of \methodName.
The trainable parameters are updated according to a loss $\mathcal{L}$, while the rest of the parameters remain unchanged. This means \methodName\ effectively leverages the large-scale pre-training of the foundation model it extends. 

The training loss $\mathcal{L}$ includes a localization term $\mathcal{L}_{loc}$ and a classification term $\mathcal{L}_{cls}$. For the localization term $\mathcal{L}_{loc}$, we regress the object centers from the final cross-modality queries output by the decoder $f_{\boldsymbol{\psi}}$, and use the $L_{1}$ loss between the predicted box center $\hat{c}$ and the ground truth $c$, similar to~\cite{zhou2019objects}.
For the classification term $\mathcal{L}_{cls}$, we compute the similarity matrix $\mathbf{\hat{Y}}$ from Equation \ref{arch_eq_2} and calculate the focal loss for each score. The final loss is:
\vspace{-2mm}
\begin{equation}\label{loss}
    \mathcal{L} = \lambda_{loc} \mathcal{L}_{loc} + \lambda_{cls} \mathcal{L}_{cls} = \lambda_{loc} \sum_{i = 1}^{l}{|\hat{c}_{i} - c_{i}|} + \lambda_{cls} \text{FocalLoss}(\mathbf{\hat{Y}}, T)
\end{equation}
where $\lambda_{loc}$ and $\lambda_{cls}$ are hyperparameters optimized using a grid search on the validation set and $T \in \{0, 1\}^{k \times (l + 1)}$ represents an optimal Hungarian matching between the $k$ predicted queries and the $l$ ground truth object instances, and the label ``no object." Refer to the finetuning strategy implemented in~\cite{Open-Grounding-Dino} for further details.

\vspace{-2mm}
\subsection{Inference}\label{inference}
\vspace{-2mm}
To predict the object count with \methodName, the image $X$, text $t$, and visual exemplars $\mathbf{B}$ are inputted to the model, outputting a similarity matrix $\mathbf{\hat{Y}} \in \mathbb{R}^{k \times (p + q)}$. The maximum score over all $p + q$ visual exemplar and text tokens is extracted for each of the $k$ queries. Maximum scores above a confidence threshold $\sigma$ are enumerated to estimate the object count. 

\section{Experiments}

\methodName\ is trained on the FSC-147~\cite{m_Ranjan-etal-CVPR21} object counting dataset training set, and then evaluated on the FSC-147 test set, and two other benchmark datasets (without any fine-tuning). We first describe the datasets, and then discuss the performance.

\subsection{Datasets \& Metrics}\label{datasets_and_metrics}
\paragraph{FSC-147~\cite{m_Ranjan-etal-CVPR21}.} FSC-147 contains 6135 images with 89 classes in the training set, 29 classes in the validation set, and 29 classes in the test set. The classes in the training, validation, and test sets do not overlap. Each image is annotated with at least three visual exemplars. For text descriptions, we use the singular forms of the class names in FSC-147-D~\cite{AminiNaieni23} with any prefixes such as ``the" removed. For example, we change ``the donuts in the donut tray" in FSC-147-D to ``donut" by removing the prefix ``the," extracting the class name ``donuts," and then singularizing it to ``donut."

\noindent {\em Corrections to FSC-147}. We make two corrections to FSC-147 and report results with and without these corrections. (1) As noted in~\cite{Liu22}, image \texttt{7171.jpg} has incorrect visual exemplars labeled. Since, unlike the model in~\cite{Liu22}, \methodName\ can input either visual exemplars or text, for this example we only provide the model with text. (2) Image \texttt{7611.jpg} has the incorrect text description ``lego" even though the lego \emph{studs} not the lego \emph{bricks} should be counted. We change the description to ``yellow lego stud" for this example.

\vspace{-2mm}
\paragraph{CARPK~\cite{Hsieh2017DroneBasedOC}. } CARPK contains images of parking lots captured by overhead drones with a training set and test set of 989 and 459 images respectively. Each image is annotated with at least two bounding boxes. We use the same two bounding boxes selected in~\cite{Liu22} as the visual exemplars for each image. We use the class name ``car" as the text description.

\vspace{-2mm}
\paragraph{CountBench~\cite{paiss2023countclip}. } CountBench contains 540 images with 2-10 objects and captions describing the image as well as the number of objects to count. We create text descriptions for a 504-image subset of CountBench, removing inappropriate images and images with links that are unavailable. We give details of how the class names are obtained from the captions accompanying each image in the Appendix.

\vspace{-2mm}
\paragraph{Metrics.} Following prior work on object counting~\cite{AminiNaieni23, Liu22, low_shot}, the Mean Absolute Error (MAE) and the Root Mean Squared Error (RMSE) are used to measure performance. We define these metrics in the Appendix. 

\subsection{Implementation}\label{sec:impl}
\paragraph{Training.} The model is trained for 30 epochs on the FSC-147 training dataset using Adam optimizer and standard augmentations. The image and text encoders, $f_{\boldsymbol{\theta_{\text{SwinT}}}}$ and $f_{\boldsymbol{\theta_{\text{TT}}}}$, are frozen during training. Full details are given in the Appendix. 

\paragraph{Inference.} At inference, each image is resized such that its shortest side length is 800 pixels, and its aspect ratio is maintained. The image is then normalized and passed to the model. The visual exemplars are passed in as bounding boxes, and the special token `` .'' is appended to the text description before providing it to the model. In the Appendix we give details of two important improvements: one to avoid double counting given self-similarity of the target object (like a butterfly~\cite{Liu22}), and the other using adaptive cropping to overcome the 900 counting quota of the the model.


\subsection{Comparison to State-of-the-art on Standard Benchmarks}
Here we show that \methodName\ achieves comparable or exceeds state-of-the-art performance for text-only open-world object counting when using only text, and exceeds the performance of all open-world object counting methods when using both visual exemplars and text on three benchmarks.
\paragraph{FSC-147~\cite{m_Ranjan-etal-CVPR21}.}

\begin{figure}[t!]
\centering
\includegraphics[width=\linewidth]{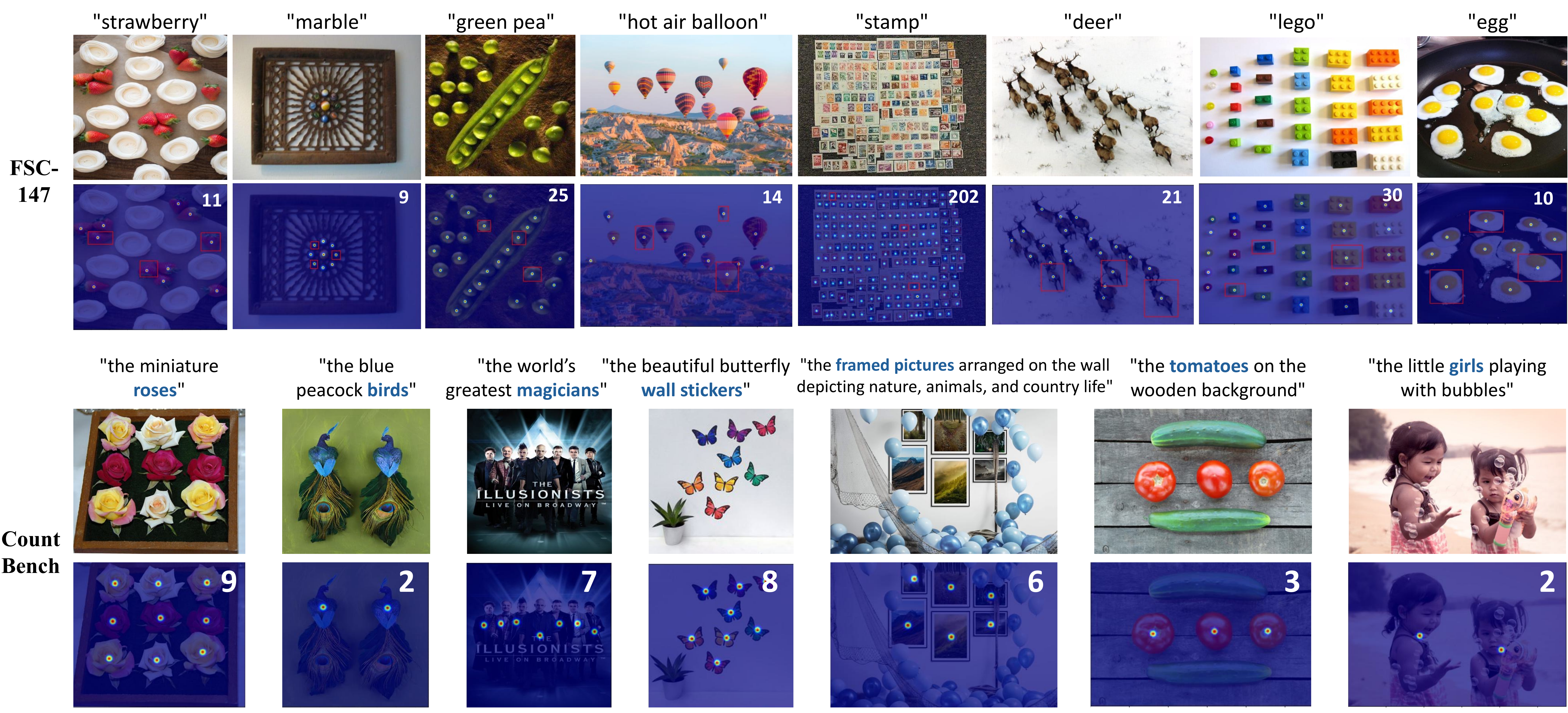}
\vspace{-6mm}
\caption{\small 
Qualitative counting results on FSC-147~\cite{m_Ranjan-etal-CVPR21} and CountBench~\cite{paiss2023countclip} using the multi-modal \methodName. The model is trained and tested on FSC-147 visual exemplars and text. Input text is written above each image, and visual exemplars are indicated by the red boxes.
On CountBench, we test the same model trained on the FSC-147 in a zero-shot way with only text (there are no visual exemplars for CountBench). Blue words indicate the subject of each caption input to the model.
In both cases, \methodName\ predicts the count in all images shown with 100\% accuracy. 
Note on the CountBench examples, the model counts the specified objects correctly when there are multiple types of objects in the image, such as the tomatoes with cucumbers, and the girls with bubbles.
Detected points are filtered with a Gaussian and plotted under the input images for visualization purposes.
}
\label{fig:fsc147_countbench}
\vspace{-2mm}
\end{figure}

\begin{table}[t!]
\begin{center}
\caption{\small \label{fsc147_table} FSC-147~\cite{m_Ranjan-etal-CVPR21} comparison with the state-of-the-art text-only and visual exemplar-only open-world counting methods. Multi-modal \methodName\ trained and tested with both visual exemplars and text achieves state-of-the-art counting accuracy for open-world object counting, beating all text-only and visual exemplar-only approaches. \methodName$_\text{txt}$ trained and tested with only text achieves comparable performance to state-of-the-art text-only counting approaches.  * = correction of erroneous GT labels, as explained in section~\ref{datasets_and_metrics}. GroundingREC~\cite{groundingrec}, DAVE$_\text{prm}$, and DAVE~\cite{dave} are concurrent work. \emph{Lower MAE and RMSE values mean more accurate results.}}
\scriptsize
\begin{NiceTabular}{l|c|c|c|c|c|c|c} 
\CodeBefore
\Body
 \hline
 \multirow{2}{*}{Method} & \multirow{2}{*}{Year} & \multirow{2}{*}{Published} & How to Specify & \multicolumn{2}{c}{Validation} & \multicolumn{2}{c}{Test} \\
  &  &  & the Class & MAE~$\downarrow$ & RMSE~$\downarrow$ & MAE~$\downarrow$ & RMSE~$\downarrow$ \\

  \hline
  Patch-selection~\cite{Xu2023ZeroshotOC} & 2023 & \cmark & Text & 26.93 & 88.63 & 22.09 & 115.17\\
  CLIP-count~\cite{Jiang2023CLIPCountTT} & 2023 & \cmark & Text & 18.79 & 61.18 & 17.78 & 106.62 \\
  VLCounter~\cite{kang2024vlcounter} & 2023 & \cmark & Text & 18.06 & 65.13 & 17.05 & 106.16 \\
  CounTX~\cite{AminiNaieni23} & 2023 & \cmark & Text & 17.10 & 65.61 & 15.88 & 106.29 \\
  CounTX$^{*}$~\cite{AminiNaieni23} & 2023 & \cmark & Text & 17.10 & 65.61 & 15.69 & 106.06 \\
  DAVE$_\text{prm}$~\cite{dave} & 2024 & \xmark & Text & 15.48 & 52.57 & 14.90 & 103.42 \\
  GroundingREC~\cite{groundingrec} & 2024 & \xmark & Text & \textbf{10.06} & 58.62 & \textbf{10.12} & 107.19 \\
  \methodName$_\text{txt}$ (ours) & 2024 & - & Text & 12.14 & 47.51 & 14.76 & 120.42 \\
  \methodName$_\text{txt}^{*}$ (ours) & 2024 & - & Text & 12.14 & \textbf{47.51} & 12.98 & \textbf{98.35} \\
  \hline
  CounTR~\cite{Liu22} & 2022 & \cmark & Visual Exemplars & 13.13 & 49.83 & 11.95 & 91.23\\
  LOCA~\cite{low_shot} & 2022 & \cmark & Visual Exemplars & 10.24 & 32.56 & 10.79 & 56.97\\
  DAVE~\cite{dave} & 2024 & \xmark & Visual Exemplars & 8.91 & 28.08 & 8.66 & 32.36 \\
  \hline
  \methodName\ (ours) & 2024 & - & Visual Exemplars \& Text & 7.10 & 26.08 & 6.75 & 43.65\\
  \textbf{\methodName$^{*}$} (ours) & \textbf{2024} & \textbf{-} & \textbf{Visual Exemplars \& Text} & \textbf{7.10} & \textbf{26.08} & \textbf{5.74} & \textbf{24.09}\\
  
 \hline
\end{NiceTabular}
\end{center}
\vspace{-6mm}
\end{table}

In Table \ref{fsc147_table}, we test \methodName\ under two settings: (1) trained and tested with only text (denoted as \methodName$_\text{txt}$), and (2) trained and tested with both 3 visual exemplars and text (denoted as \methodName). \methodName\ trained and tested with both visual exemplars and text sets a new state-of-the-art for counting accuracy on FSC-147, achieving significantly lower counting errors than all prior approaches to open-world object counting. Training with only text achieves comparable counting accuracy to state-of-the-art text-only open-world object counting methods. The concurrent method GroundingREC~\cite{groundingrec} achieves slightly lower mean absolute error values than \methodName$_\text{txt}$, while \methodName$_\text{txt}$\ achieves lower root mean squared error values. The results for GroundingREC and \methodName$_\text{txt}$ are likely close to each other since both methods leverage the pretrained GroundingDINO~\cite{liu2023grounding} foundation model. However, unlike GroundingREC, \methodName\ can fuse information from both text and visual exemplars instead of using only text, enabling a significant improvement.

In Figure~\ref{fig:fsc147_countbench}, we give qualitative examples of the detections that \methodName\ outputs given both visual exemplars and text from the FSC-147 test set. Note how in the first image, \methodName\ only counts the strawberries and not the white cookies. Prior work has shown that visual exemplar-only methods struggle to count only one category of object when there are repeating instances from multiple categories in an image~\cite{dave}. \methodName\ handles this issue very well in this example by leveraging the generalization capabilities of the pretrained vision-language model GroundingDINO~\cite{liu2023grounding}.

\begin{table}[t!]
\begin{center}
\caption{\small \label{carpk_countbench_table} 
Comparison with state-of-the-art open-world counting methods.
\textbf{(top)} On CARPK~\cite{Hsieh2017DroneBasedOC}, we compare with text-only and visual exemplar-only methods. \methodName, trained with both visual exemplars and text on FSC-147~\cite{m_Ranjan-etal-CVPR21}, achieves lower error values than all text-only and visual exemplar-only methods, without being trained on any images in CARPK, using either text-only or both text and two visual exemplars at inference.
\textbf{(bottom)} On CountBench~\cite{paiss2023countclip}, we compare with currently the best 
publicly available text-only open-world counting method, CounTX~\cite{AminiNaieni23}. \methodName\ (trained on both visual exemplars and text), given only text and zero-shot, achieves significantly lower errors than CounTX. 
Note, CountBench does not provide visual exemplars.
}
\scriptsize
\begin{NiceTabular}{c|l|c|c|c|c|c|c} 
\CodeBefore
\Body
 \hline
 \multirow{2}{*}{Dataset} & \multirow{2}{*}{Method} & \multirow{2}{*}{Year} & \multirow{2}{*}{Published} & How to Specify & \multirow{2}{*}{Fine-tuned} & \multicolumn{2}{c}{Test} \\
 &  &  &  & the Class & & MAE~$\downarrow$ & RMSE~$\downarrow$ \\
  \hline
  \multirow{8}{*}{CARPK}
  & CLIP-count~\cite{Jiang2023CLIPCountTT} & 2023 & \cmark & Text & \xmark & 11.96 & 16.61 \\
  & CounTX~\cite{AminiNaieni23} & 2023 & \cmark & Text & \cmark & 8.13 & 10.87 \\
  & VLCounter~\cite{kang2024vlcounter} & 2023 & \cmark & Text & \xmark & 6.46 & 8.68 \\
  & \textbf{\methodName}~(ours) & \textbf{2024} & \textbf{-} & \textbf{Text} & \xmark & \textbf{3.83} & \textbf{5.41} \\
  & LOCA~\cite{low_shot} & 2022 & \cmark & Visual Exemplars & \xmark & 9.97 & 12.51 \\
  & CounTR~\cite{Liu22} & 2022 & \cmark & Visual Exemplars & \cmark & 5.75 & 7.45\\
  & SAFECount~\cite{You_2023_WACV} & 2022 & \cmark & Visual Exemplars & \cmark & 5.33 & 7.04\\
  & \textbf{\methodName}~(ours) & \textbf{2024} & \textbf{-} & \textbf{Visual Exemplars \& Text} & \xmark & \textbf{3.68} & \textbf{5.17}\\ 
  \hline
  \multirow{2}{*}{CountBench}
  & CounTX~\cite{AminiNaieni23} & 2023 & \cmark & Text & \xmark & 6.64 & 15.75 \\
  & \textbf{\methodName}~(ours) & \textbf{2024} & \textbf{-} & \textbf{Text} & \xmark & \textbf{0.86} & \textbf{3.1}\\
 \hline
\end{NiceTabular}
\end{center}
\vspace{-4mm}
\end{table}

\paragraph{CARPK~\cite{Hsieh2017DroneBasedOC}.} To test cross-dataset generalization, \methodName\ is  trained on FSC-147~\cite{m_Ranjan-etal-CVPR21}, and tested on the CARPK car counting dataset zero-shot, without being trained on any images in CARPK. 
In Table~\ref{carpk_countbench_table}, \methodName\ is trained on the FSC-147~\cite{m_Ranjan-etal-CVPR21} training set with both visual exemplars and text, and tested on the CARPK car counting dataset under two settings: (1) using only the text input ``car", and (2) using both the text input ``car" and the same two visual exemplars as~\cite{Liu22}. Under both settings, \methodName\ achieves state-of-the-art accuracy on CARPK for all open-world object counting methods without being trained on any images in CARPK, achieving lower counting errors than methods like CounTR~\cite{Liu22} and SAFECount~\cite{You_2023_WACV} that were fine-tuned on CARPK. 

\paragraph{CountBench~\cite{paiss2023countclip}.}  We train \methodName\ on FSC-147, which has at least seven objects in each training image, and evaluate its generalization to  counting low numbers of objects in the CountBench test set zero-shot.
In Table~\ref{carpk_countbench_table}, we compare \methodName's performance on counting low numbers of objects (2--10) to CounTX~\cite{AminiNaieni23}, 
currently the best
(according to performance on FSC-147~\cite{m_Ranjan-etal-CVPR21}) publicly available pre-trained open-world text-specified object counting methods. For this experiment, \methodName\, trained with both visual exemplars and text on FSC-147, is tested on CountBench zero shot given only text. Because CountBench contains long captions that describe more than the object to count, we only threshold text token similarity scores corresponding to the subject of each caption. \methodName\ achieves significantly better performance than CounTX on this dataset. 
In Figure~\ref{fig:fsc147_countbench}, we show qualitative examples of the detections output by \methodName. The subject of each caption is shown with yellow text. 

\begin{table}[t!]
\begin{center}
\caption{\small \label{uni_v_multi_modal} Ablation study I: \methodName\ trained and tested with text only, visual exemplars only, and text and visual exemplars together on FSC-147~\cite{m_Ranjan-etal-CVPR21}. Multi-modal \methodName\ trained and tested with both text and visual exemplars achieves the lowest counting errors.}
\scriptsize
\begin{NiceTabular}{l|c|c|c|c} 
\CodeBefore
\Body
 \hline
 \multirow{2}{*}{Training and testing setting} & \multicolumn{2}{c}{Val} & \multicolumn{2}{c}{Test} \\
  &  MAE~$\downarrow$ & RMSE~$\downarrow$ & MAE~$\downarrow$ & RMSE~$\downarrow$ \\
  \hline 
  \methodName\ (Text) & 12.14 & 47.51 & 12.98 & 98.35 \\
  \methodName\ (Visual Exemplars) & 7.46 & 29.54 & 8.31 & 91.05 \\
  \textbf{\methodName\ (Text \& Visual Exemplars)} & \textbf{7.10} & \textbf{26.08} & \textbf{5.74} & \textbf{24.09}\\
 \hline
\end{NiceTabular}
\end{center}
\vspace{-6mm}
\end{table}

\subsection{Ablation Study}
\paragraph{Uni-Modal vs.\ Multi-Modal Training.}
In Table~\ref{uni_v_multi_modal}, we compare \methodName's performance using different training and inference procedures on FSC-147~\cite{m_Ranjan-etal-CVPR21}. Training on text only and testing with text only achieves performance comparable to state-of-the-art counting accuracy for text-only approaches, demonstrating the superiority of the GroundingDINO~\cite{liu2023grounding} architecture that we leverage. Training  with visual exemplars only and testing with visual exemplars only results in state-of-the-art performance on two out of four of the metrics (mean absolute errors on both the validation and test sets) for visual exemplar-only approaches. This is surprising given that GroundingDINO was pretrained to relate text to images not visual exemplars to images. Despite this, \methodName\ performs remarkably well in this setting. Multi-modal training and testing with both visual exemplars and text beats both uni-modal approaches and sets a new state-of-the-art for open-world object counting. This ablation study shows that the visual exemplars provide more information than the text in FSC-147 as the performance with visual exemplars only is significantly better than the performance with text only. It also demonstrates that multi-modal training and inference is the superior strategy as it allows \methodName\ to take advantage of two sources of information about the object instead of one. In Table \ref{inference_ablation} in the Appendix, we additionally include an ablation study showing the influence of our proposed SAM Test-time normalization and adaptive cropping strategies.

\begin{figure}[t!]
\centering
\includegraphics[width=\linewidth]{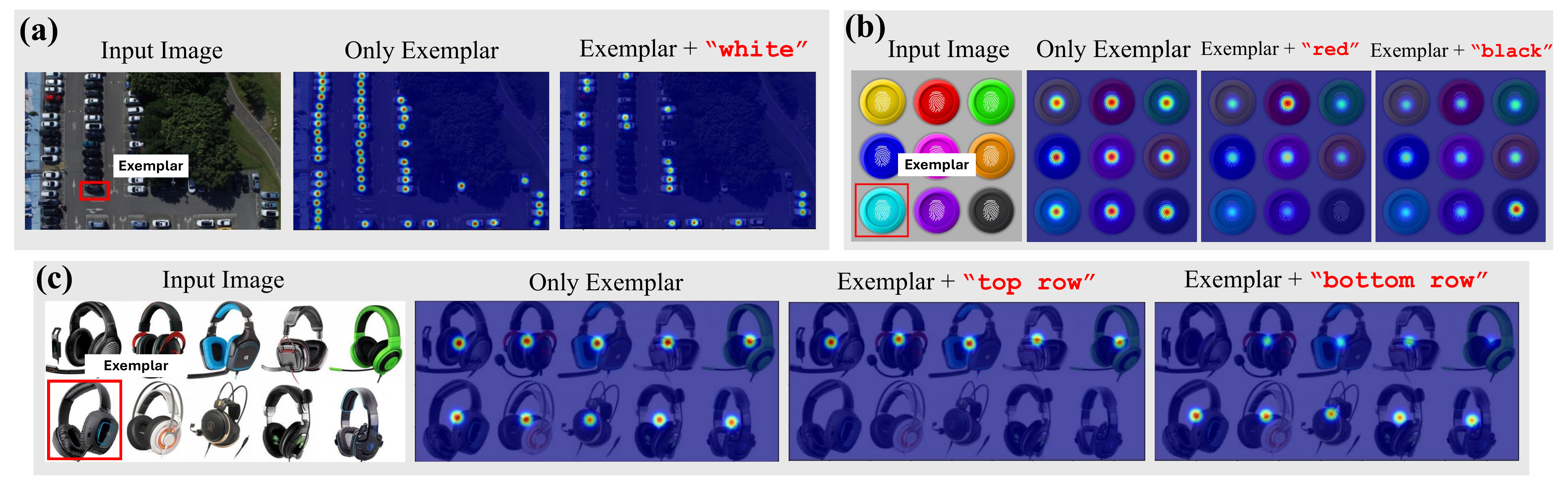}
\vspace{-2mm}
\caption{\small Studying visual exemplar and text interactions. We plot the confidence scores of the instances for each image. In (a) and (b) we show we can specify shape with the exemplar and modify color with text. In (c) we show we can specify spatial location with text, and shape with the exemplar.}
\label{interaction_study_fig}
\vspace{-5mm}
\end{figure}

\vspace{-2mm}
\subsection{Language and Exemplar Interactions}
\vspace{-2mm}
Up to this point we have used the text and visual exemplar prompts to specify the target object in a complementary manner; for example giving a visual exemplar of a `strawberry' with the text `strawberry'. It has been seen that the counting performance with prompts in both modalities is, in general, equal or superior to text alone. In this section we investigate qualitatively the case where the text refines or filters the visual information provided by the exemplars. For example, where the visual exemplar is car, but the text specifies the color, and only cars of that color are counted. 

In this study, unlike before, we freeze the feature enhancer in addition to the image and text encoders and finetune the rest of the model on FSC-147~\cite{m_Ranjan-etal-CVPR21}. We find that freezing the feature enhancer is necessary for many of these interactions to emerge. Once trained, the new model can use the text to filter instances picked out by the exemplar, and the exemplar can increase the confidence when it reinforces the text. In Figure \ref{interaction_study_fig} we show several examples of the interactions observed.

\vspace{-2mm}
\section{Conclusion \& Future Work}
\vspace{-2mm}
We have extended the generality of open-world counting by introducing a model that can accept visual exemplars or text descriptions or both as prompts to specify the target object to count. The complementarity of these prompts in turn leads to improved counting performance.
There are three research directions that naturally follow on from this work: (i) the performance could probably be further improved by training on larger scale datasets, for example using synthetic data as demonstrated recently for counting~\cite{knobel2023selfcollages}; (ii) a larger training set would enable a thorough investigation of freezing more of the GroundingDINO model when adding our new visual exemplar modules; and finally, (iii) the model does not currently predict the errors of its counting. We discuss this point in the Limitations in the Appendix.

\subsubsection*{Acknowledgement}
The authors would like to thank Shilong Liu for his extensive support of GroundingDINO~\cite{liu2023grounding}, Zechen Bai for his extensive support of Patch CLIP introduced in~\cite{patchclip}, and Kiana Amini-Naieni for her help in labeling the CountBench~\cite{paiss2023countclip} images. We would also like to thank Oishi Deb, Abhishek Dutta, Horace Lee, Orest Kupyn, Vladimir Lashin, and Paul Engstler for providing detailed feedback on the CountGD app. This research is funded by an AWS Studentship, the Reuben Foundation, the AIMS CDT program at the University of Oxford, EPSRC Programme Grant VisualAI EP/T028572/1, and a Royal Society Research Professorship RP \textbackslash R1 \textbackslash 191132.

\clearpage
\bibliography{bib/longstrings, bib/vgg_local, bib/other}

\clearpage
\appendix
{\LARGE \textbf{Appendix}}
\section{Definition of Metrics}
We use the Mean Absolute Error (MAE) and the Root Mean Squared Error (RMSE) to measure performance. 
They are defined as:
\begin{equation}\label{err_forms}
    \text{MAE} = \frac{1}{N}\sum_{i = 1}^{N}|\hat{y}_{i} - y_{i}|,\quad\text{ }\text{RMSE} = \sqrt{\frac{1}{N}\sum_{i = 1}^{N}(\hat{y}_{i} - y_{i})^2}
\end{equation}
where $N$ is the number of test images, $\hat{y}_{i}$ is the predicted count, and $y_{i}$ is the ground truth count for image $X_{i}$.
For both MAE and RMSE, a lower value indicates a better performance.

\section{Additional Dataset Details}

\paragraph{CountBench~\cite{paiss2023countclip}}
Here we explain how the descriptions and keywords for CountBench were constructed. Unlike the original CountBench captions, our text descriptions include the object to count without revealing the number of objects. For example, the caption ``background photo of three light bulbs" in CountBench is replaced with ``the light bulbs," which describes the object to count (the light bulbs) without giving away that there are three in the image. Because some descriptions include information about other objects in the image, we add keywords that indicate the subject in the caption. For example, the caption ``the children standing on a bench at an outdoors party" includes the keyword ``children" to indicate that the children, not the bench, should be counted. Providing keywords is necessary since \methodName\ has been pretrained on visual grounding data and will count both the children and the bench they are sitting on as a result. To ensure only the children are counted, text token similarity scores from the keyword ``children" are thresholded to estimate the count. Because CountBench contains very few objects, we do not use visual exemplars and provide \methodName\ with just the text description.

\section{Additional Implementation Details}

\paragraph{Architecture.}

We provide additional architectural details here. The image encoder, $f_{\boldsymbol{\theta_{\text{SwinT}}}}$, is a Swin-B transformer with corresponding patch sizes $8 \times 8$, $16 \times 16$, and $32 \times 32$ and final embedding dimensions of $192$, $384$, and $768$ respectively. To get the visual exemplar tokens, the spatial feature maps from $f_{\boldsymbol{\theta_{\text{SwinT}}}}(\mathbf{X})$ are first upsampled to the same height and width as the largest one with patch size $8 \times 8$. The upsampled feature maps are concatenated along the channel dimension and projected to 256 dimensions with a separate $1\times 1$ convolution. RoIAlign is then applied to extract features corresponding to the exemplar regions, 

\paragraph{Training.}

Each training image is first horizontally flipped with a probability of $50\%$. Next, with a probability of $50\%$, either the minimum side length of the image is resized to a side length in $\{480, 512, 544, 576, 608, 640, 672, 704, 736, 768, 800\}$ such that the aspect ratio of the image is maintained or the image is first randomly cropped such that the minimum side length is in the range $[384, 600]$ and then resized as mentioned before. After this, the image is normalized and passed through the model. Following~\cite{liu2023grounding}, all the classes in the FSC-147 training set are concatenated into a single caption with `` .'' separating each class name. Visual exemplar tokens are appended to the end of the text tokens associated with their class name. Self-attention masks are constructed such that the text tokens attend to each other as well as to the visual exemplars that are associated with their class name. Self-attention is not applied between unrelated class names and visual exemplars. The model is optimized with the Adam Optimizer with a weight decay set to $10^{-4}$ and an initial learning rate set to $1 \times 10^{-4}$ that reduces by a factor of ten every ten epochs. $\lambda_{loc}$ is set to 1 and $\lambda_{cls}$ is set to 5 in Equation \ref{loss}. These scale factors are also used in the Hungarian Matching Cost for matching ground truth points to predicted points. The confidence threshold $\sigma$ is set to 0.23. Hyperparameters are set using default values provided by~\cite{Open-Grounding-Dino} with the exception of selecting $\lambda_{loc}$, $\lambda_{cls}$, and the confidence threshold $\sigma$ using a sparse grid search optimizing the mean absolute counting error on the validation set. Specifically $(\lambda_{loc}, \lambda_{cls}) \in \{1, 2.5, 5\} \times \{1, 2.5, 5\}$ and $\sigma \in \{0.14, 0.17, 0.2, 0.23, 0.26\}$ are tested. The model is trained for 30 epochs with early stopping with respect to the mean absolute counting error on the validation set with no SAM TT-Norm or adaptive cropping applied. We train the multi-modal model that beats all prior methods for open-world object counting using both visual exemplars and text. We train the text-only model that beats all text-only approaches to open-world object counting on only text. Finally, we additionally freeze the feature enhancer for the exemplar and text interaction study.

\paragraph{Training Resources.}
Our model is trained on 1 Nvidia A6000 GPU with 48GB of graphic memory. A full training takes about 1 day.

\section{Additional Inference Details}

\begin{figure}
\centering
\begin{subfigure}[h]{0.35\linewidth}
\includegraphics[width=\linewidth]{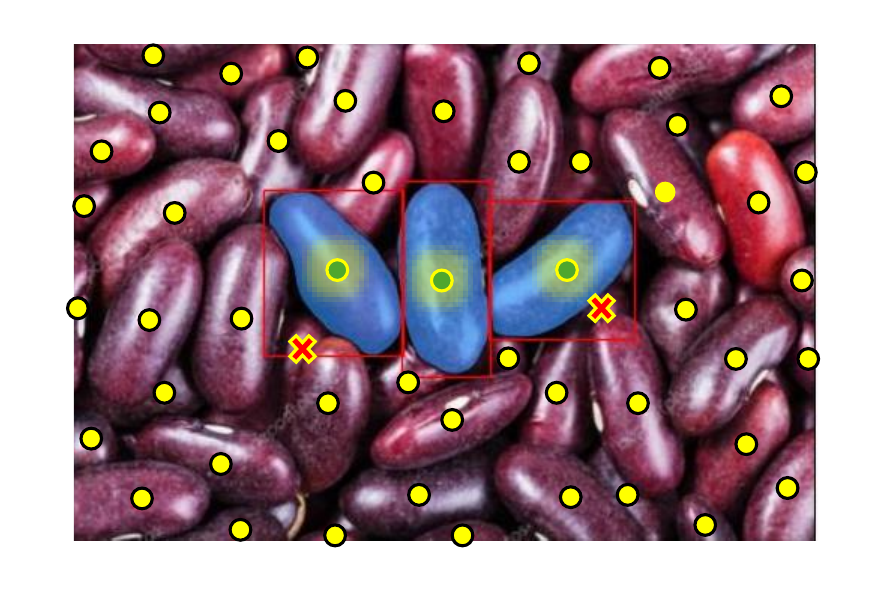}
\caption{CounTR~\cite{Liu22} incorrectly detects self-similarity.}
\end{subfigure}
\hspace{2pt}
\begin{subfigure}[h]{0.35\linewidth}
\includegraphics[width=\linewidth]{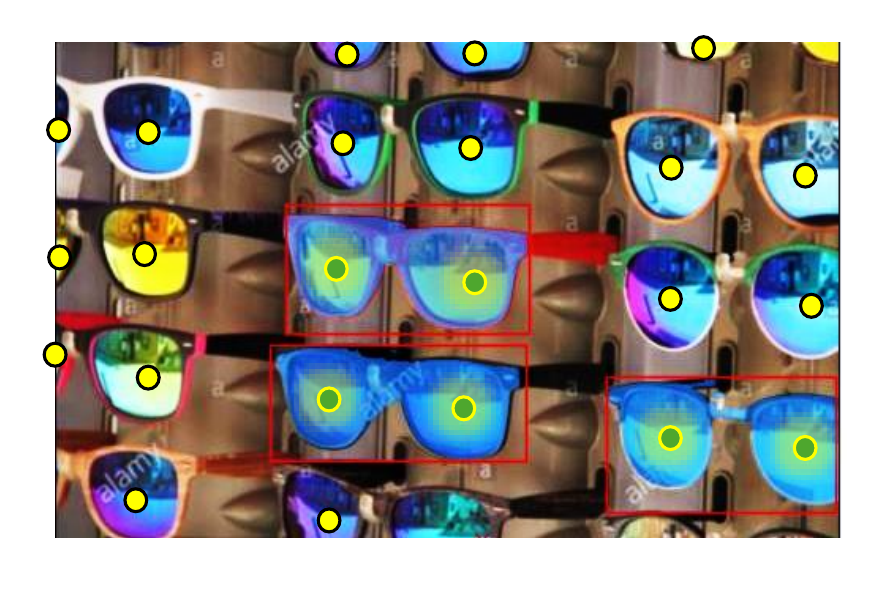}
\caption{SAM TT-Norm correctly detects self-similarity.}
\end{subfigure}%
\caption{In example (a), the TT-Norm presented in~\cite{Liu22} would incorrectly detect self-similarity since multiple correctly detected instances, denoted by red `{\color{red}$\times$}'s, fall within the exemplar regions, indicated by the red boxes. However, the blue segmentation masks output by SAM only contain one detected instance per mask, so the SAM TT-Norm correctly does not detect self-similarity. (b) shows a case where the SAM TT-Norm correctly detects self-similarity and divides the estimated count by 2.}\label{sam_tt_norm}
\end{figure}

\paragraph{Avoiding double counting.}

One of the common problems for counting models is handling self-similarity, when an object is intrinsically repetitive. For example, sunglasses and butterflies exhibit self-similarity. In these cases, counting methods tend to double count, detecting each self-similar component. CounTR~\cite{Liu22} has tried to address this by dividing the estimated count by the average count in the visual exemplar regions. We observe that this approach (referred to as the ``TT-Norm'') fails in cluttered scenes, where visual exemplar bounding boxes encapsulate more than just one instance of the object. This causes the counting model to detect self-similarities when they are not present. We show an example of this in Figure \ref{sam_tt_norm} (a).

To address this, we propose to use segmentation masks instead of bounding boxes to more accurately check if the counting model detects more than one instance on an object inside a visual exemplar. To obtain the segmentation masks we use the visual exemplars as box prompts to the Segment Anything Model (SAM)~\cite{sam}. This approach avoids the issue faced by CounTR's TT-Norm since we do not check instances outside the object's boundary, even if these instances fall within the visual exemplar regions. Figure \ref{sam_tt_norm} shows how our approach compares to CounTR's TT-Norm in such cases.

\paragraph{Adaptive Cropping.}
We address the problem that \methodName\ can only output at most 900 queries at a time through adaptive cropping. If \methodName\ detects 900 objects, we crop the input image into multiple overlapping pieces and pass each cropped piece back to the model. The crop width and height are calculated as 4 times the average exemplar width and height. This approximately upper bounds the number of objects that can appear inside the crop window. The overlap width and height are determined to be 1.25 times the average exemplar width and height to approximately ensure each object instance appears fully in at least one crop. If visual exemplars are not provided, the image is cropped into four equal pieces. To obtain the final count, the number of detected instances in each crop window are added together while averaging the predicted count in overlapping regions. 

\paragraph{Ablation Study.}

\begin{table}[t!]
\begin{center}
\caption{\small \label{inference_ablation} Ablation study II: \methodName\ is tested with different inference procedures on FSC-147~\cite{m_Ranjan-etal-CVPR21}. TT-Norm refers to test-time normalization. Correction (1) refers to using only text for image \texttt{7171.jpg}, and Correction (2) refers to correcting the incorrect text description for image \texttt{7611.jpg} from ``lego'' to ``yellow lego stud.''}
\scriptsize
\begin{NiceTabular}{c|c|c|c|c|c|c|c} 
\CodeBefore
\Body
 \hline
 \multirow{2}{*}{SAM TT-Norm} & \multirow{2}{*}{Adaptive Cropping} & \multirow{2}{*}{Correction (1)} & \multirow{2}{*}{Correction (2)} & \multicolumn{2}{c}{Val} & \multicolumn{2}{c}{Test} \\
  & & & & MAE~$\downarrow$ & RMSE~$\downarrow$ & MAE~$\downarrow$ & RMSE~$\downarrow$ \\
  \hline 
  \xmark & \xmark & \xmark & \xmark & 8.69 & 43.89 & 10.92 & 99.58 \\
  \cmark & \xmark & \xmark & \xmark & 7.99 & 42.23 & 9.62 & 98.90 \\
  \cmark & \cmark & \xmark & \xmark & 7.10 & 26.08 & 6.75	& 43.65 \\
  \cmark & \cmark & \cmark & \xmark & 7.10 & 26.08 & \textbf{5.70} & \textbf{24.04} \\
  \cmark & \cmark & \cmark & \cmark & \textbf{7.10} & \textbf{26.08} & 5.74 & 24.09 \\
  \hline
\end{NiceTabular}
\end{center}
\vspace{-4mm}
\end{table}

In Table~\ref{inference_ablation}, we test the influence of the SAM TT-Norm, Adaptive Cropping, and our two corrections to the FSC-147 annotations on the multi-modal \methodName\ model. The SAM TT-Norm provides minor improvements. This is because only a small number of classes in FSC-147 exhibit self-similarity. The adaptive cropping provides significant improvements with respect to the RMSE and minor improvements with respect to the MAE. This is because the adaptive cropping is specifically for handling high counts of objects ($\geq 900$), and the RMSE is particularly sensitive to errors from these outliers. Correction (1), only using text for image \texttt{7171.jpg}, has a significant influence on the RMSE as using the incorrectly annotated visual exemplars for this example causes \methodName\ to correctly count the lego studs identified by the exemplars, not the lego bricks, and there are a high number of studs and a small number of bricks. Providing only the text ``lego" and discarding the erroneous visual exemplars fixes this issue. Correction (2), correcting the text description for \texttt{7611.jpg}, has no significant influence on the multi-modal model.

\section{Additional Qualitative results}

\begin{figure}[h!]
\centering
\includegraphics[width=\linewidth]{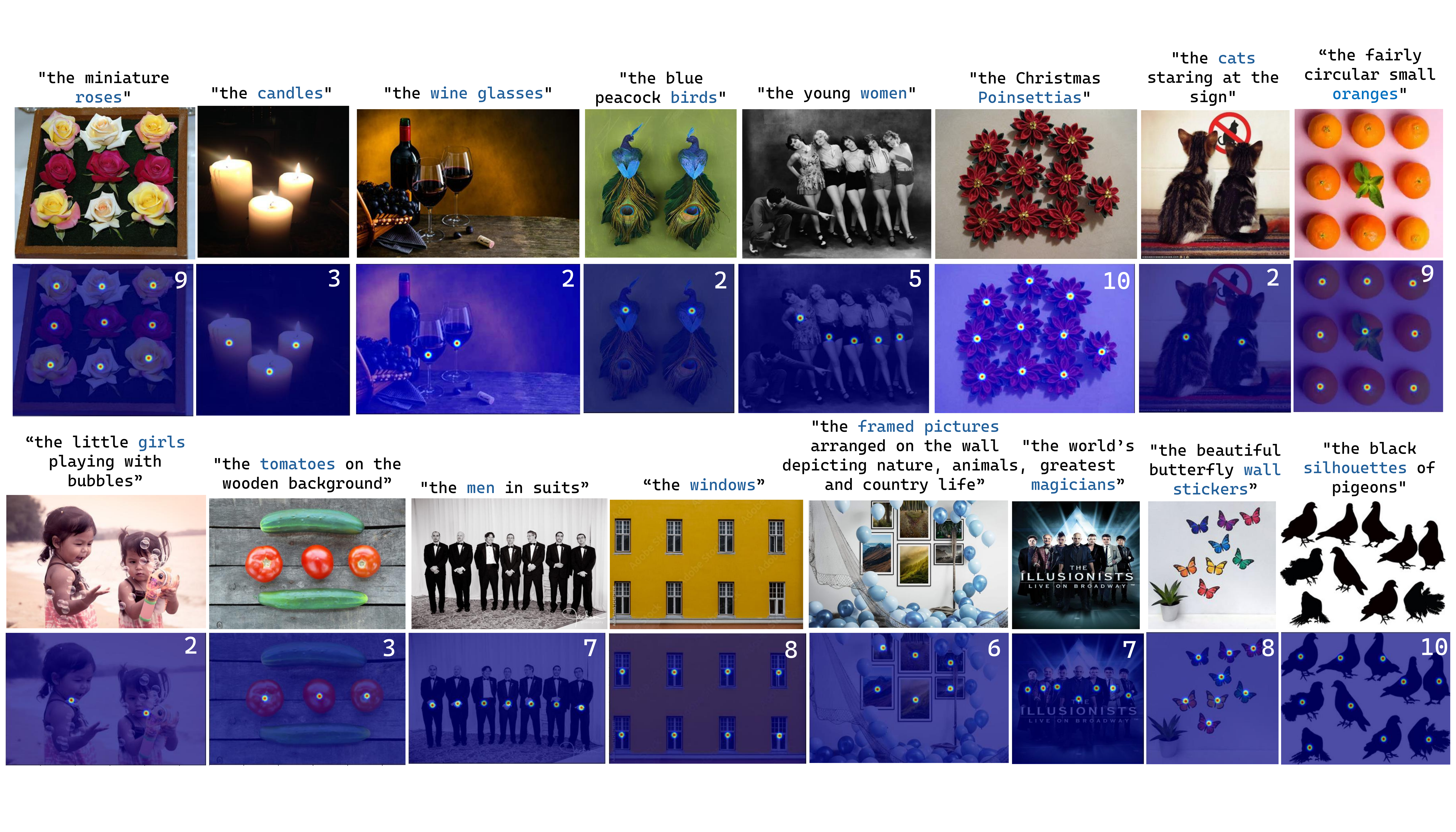}

\caption{
CountBench~\cite{paiss2023countclip} counting examples using the multi-modal \methodName. The model is trained on visual exemplars and text from FSC-147~\cite{m_Ranjan-etal-CVPR21} and tested zero-shot with text only on CountBench. Blue letters indicate the subject of each caption input to the model. Detected points are filtered with a Gaussian and plotted under input images for visualization purposes. \methodName\ predicts the count in all the images shown with 100\% accuracy. {\bf Note} how in the top row \methodName\ correctly only counts the women, not the men, in ``the young women" example, and only the alive cats (not the one painted on the wall) in ``the cats starting at a wall". In the bottom row, the model also correctly does {\em not} count the repeated bubbles near the little girls or the multiple balloons.
}
\label{add_countbench_fig}
\end{figure}

\begin{figure}[h!]
\centering
\includegraphics[width=\linewidth]{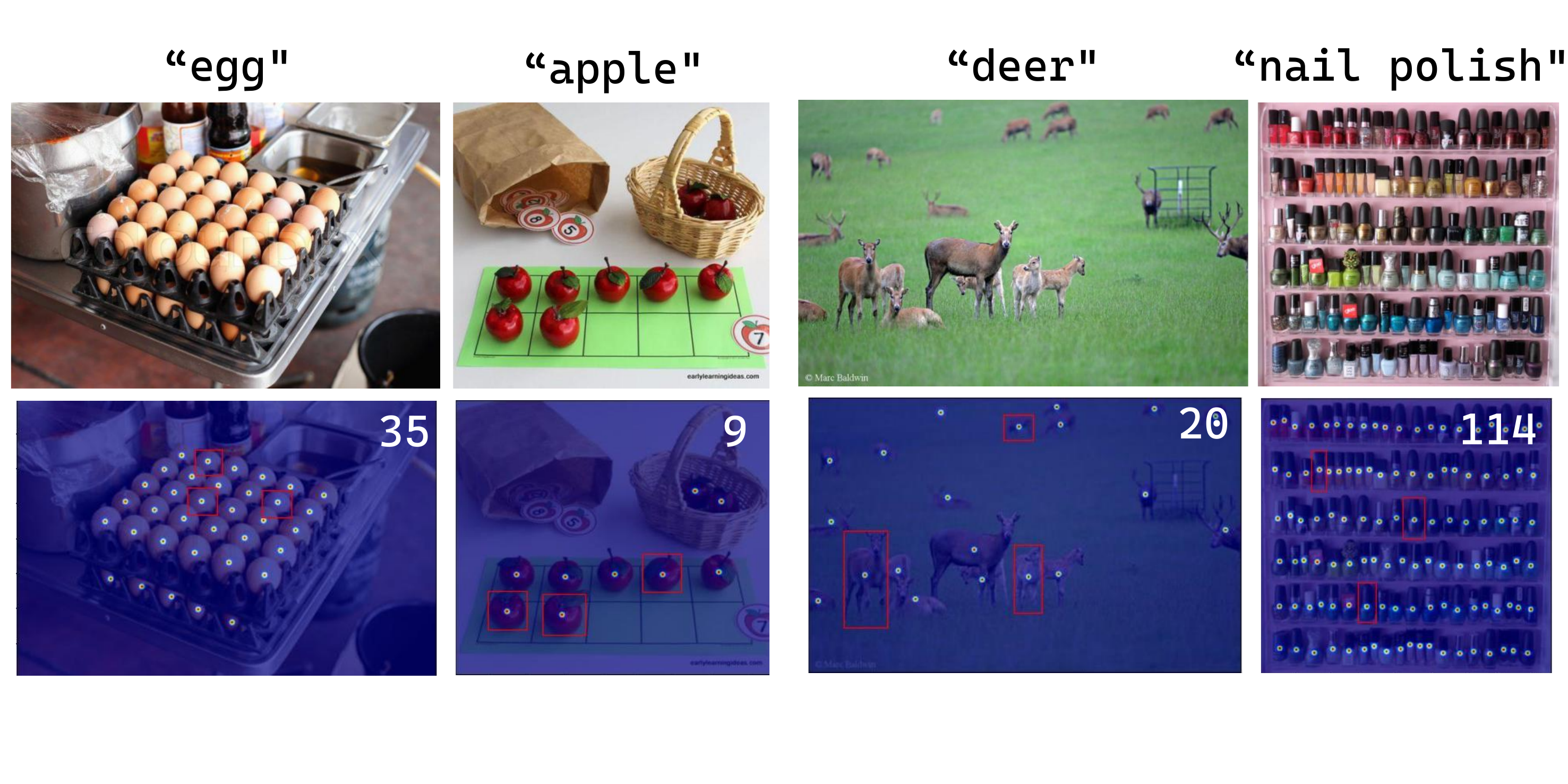}

\caption{Additional qualitative examples showing CountGD's performance on the FSC-147 test set. In these examples, CountGD predicts the count with 100 \% accuracy.}
\label{add_fsc147_fig}
\end{figure}

\section{Limitations}
Somewhat oddly, given the quantitative nature of the task, recent papers on counting do not generally include uncertainty or error bars. This is probably because the main drive at the moment is the move towards more open-world methods.
Some early counting papers did include counting error estimation, e.g.~\cite{Arteta16,AminiNaieni23,Jiang2023CLIPCountTT}.
We recognize that this is an important limitation to address, and given that our method relies on a threshold, we are in a good position to push this direction in future research.

\section{Broader Impacts}
In general, object counting is an important task with many real-world applications.
A strong counting model has positive impacts on various domains such as agriculture, satellite images, microscopy and medical images. 
However, there are also possible negative impacts, like violating privacy in human counting for surveillance cameras, or being used for military applications.
This paper focuses on building a more accurate and flexible counting model. 
The broader impacts in real-world scenarios should be considered carefully before deploying the model.

\end{document}